\documentclass{article} 
\usepackage{preprint,times}

\usepackage{amsmath,amsfonts,bm}

\def\eqref#1{equation~\ref{#1}}

\def\1{\bm{1}}

\DeclareMathAlphabet{\mathsfit}{\encodingdefault}{\sfdefault}{m}{sl}
\SetMathAlphabet{\mathsfit}{bold}{\encodingdefault}{\sfdefault}{bx}{n}

\def\gR{{\mathcal{R}}}

\def\gU{{\mathcal{U}}}

\newcommand{\E}{\mathbb{E}}

\usepackage[colorlinks=true,linkcolor=blue,citecolor=blue,urlcolor=blue]{hyperref}
\usepackage{url}
\usepackage{graphicx}
\usepackage{subcaption}
\usepackage{amsthm}
\usepackage{pifont}
\usepackage{booktabs}
\usepackage{arydshln}
\usepackage{amsmath}
\usepackage{amssymb}
\usepackage{algorithm}
\usepackage{algpseudocode}
\usepackage{multirow}
\usepackage{natbib}

\newtheorem{theorem}{Theorem}[section]
\newtheorem{lemma}[theorem]{Lemma}
\newtheorem{assumption}{Assumption}
\theoremstyle{definition}
\newtheorem{definition}{Definition}
\theoremstyle{remark}

\DeclareMathOperator{\err}{err}
\DeclareMathOperator{\Pbb}{\mathbb{P}}
\newtheorem{proposition}[theorem]{Proposition}

\newcommand{\ourapproach}{Saber}

\iclrfinalcopy
\title{Saber: An Efficient Sampling with Adaptive Acceleration and Backtracking Enhanced Remasking for Diffusion Language Model}

\title{Saber: An Efficient Sampling with Adaptive Acceleration and Backtracking Enhanced Remasking for Diffusion Language Model}

\author{Yihong Dong$^{1,2}$, Zhaoyu Ma$^{1}$, Xue Jiang$^{1,2}$, Zhiyuan Fan$^{1}$, Jiaru Qian$^{1}$, Yongmin Li$^{1}$, \\ \textbf{Jianha Xiao$^{1}$, Zhi Jin$^{1}$, Rongyu Cao$^{2}$, Binhua Li$^{2}$, Fei Huang$^{2}$, Yongbin Li$^{2}$, Ge Li$^{1}$}\\
$^1$ School of Computer Science, Peking University \\$^2$ Tongyi Lab, Alibaba Group\\
\texttt{\{dongyh, mazhaoyu\}@stu.pku.edu.cn} \quad \texttt{lige@pku.edu.cn}\\\
}

\begin{document}

\maketitle

\begin{abstract}
Diffusion language models (DLMs) are emerging as a compelling alternative to the dominant autoregressive paradigm, offering inherent advantages in parallel generation and bidirectional context modeling.  
However, for the tasks with strict structural constraints such as code generation, DLMs face a critical trade-off between inference speed and output quality, where accelerating generation by reducing sampling steps often leads to catastrophic performance collapse.
We find that the fundamental reasons are: 1) the generation difficulty is non-uniform in the structured sequence decoding steps, making DLM's static acceleration strategy suboptimal; 2) the context of tokens generated by DLM evolves continuously, causing early high-confidence predictions to turn into irreversible errors.
In this paper, we introduce efficient \textbf{S}ampling with \textbf{A}daptive acceleration and \textbf{B}acktracking \textbf{E}nhanced \textbf{R}emasking (i.e., \textbf{Saber}), a novel training-free sampling algorithm for DLMs that first achieves both better inference speed and output quality in code generation. Saber dynamically adjusts the number of tokens unmasked per step based on the model's evolving confidence, and utilizes a backtracking mechanism to revert tokens whose confidence drops as new context emerges, with its effectiveness supported by theoretical analysis.
Extensive experiments on multiple mainstream code generation benchmarks show that Saber boosts Pass@1 accuracy by an average of 1.9\% over mainstream DLM sampling methods, while achieving an average 251.4\% inference speedup. By leveraging the inherent advantages of DLMs, our work significantly narrows the performance gap with autoregressive models in code generation.\footnote{Our code is available at \url{https://github.com/zhaoyMa/Saber}.}
\end{abstract}

\section{Introduction}
Diffusion language models (DLMs) have emerged as a promising non-autoregressive alternative in the field of natural language processing (NLP), with inherent advantages in parallel decoding and bidirectional context modeling through iterative denoising processes \cite{d3pm,ou2025your,llada,Ye2025Dream7D}. Unlike existing autoregressive models (ARMs) that generate text left-to-right \cite{Radford2018ImprovingLU,Radford2019LanguageMA,Brown2020LanguageMA,Touvron2023LLaMAOA}, DLMs can simultaneously update multiple token positions by progressively unmasking the generation sequence, enabling global planning and iterative refinement \cite{ye2025beyond,Gong2025DiffuCoderUA}. This paradigm is especially compelling for structured generation tasks like code generation. 

\begin{figure*}
    \centering
    \includegraphics[width=0.98\linewidth]{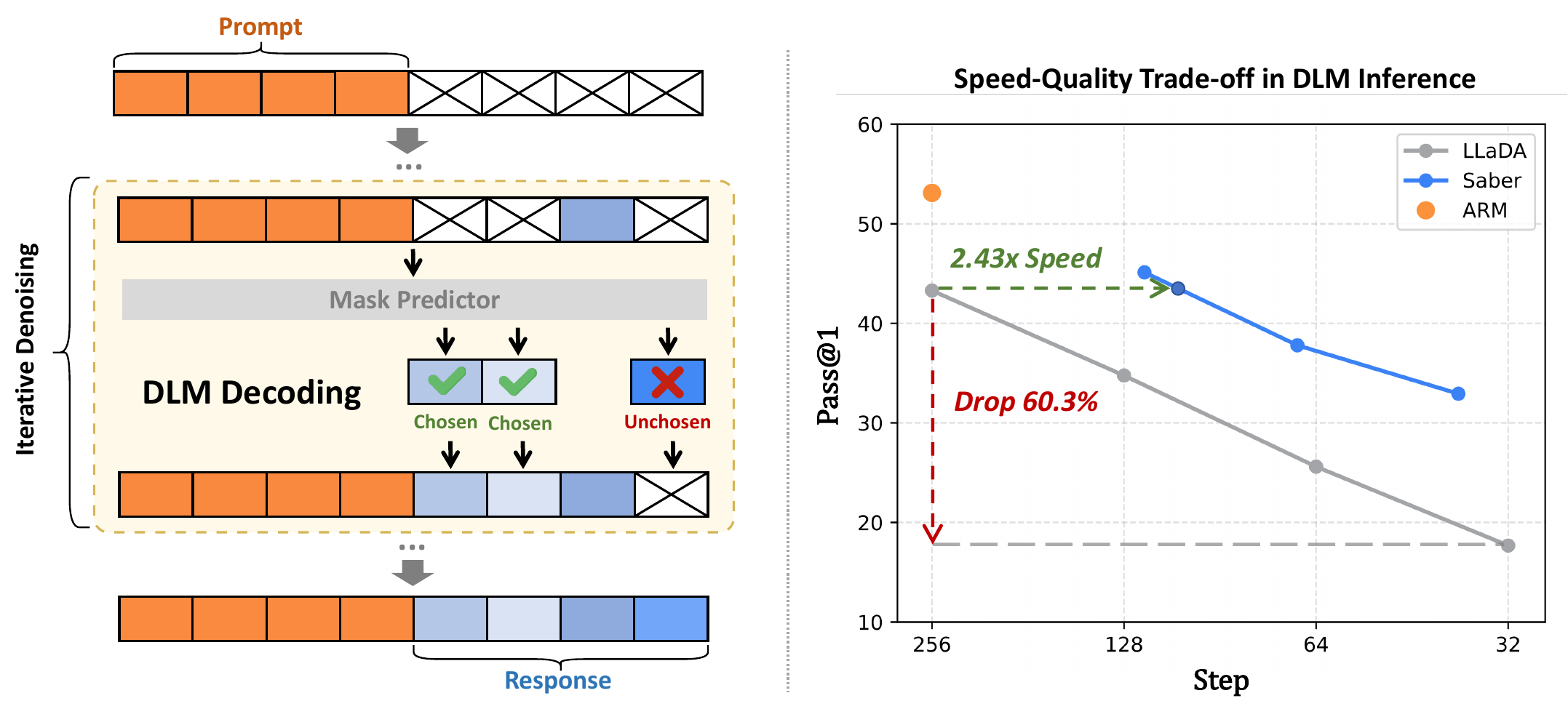}
    \caption{Left: Illustration of DLM Sampling. Right: The trade-off of DLM Sampling between inference speed and output quality on a representative benchmark (HumanEval).}
    \label{fig:intro}
\end{figure*}

Despite the potential advantages, DLMs still lag behind ARMs in practical performance, especially for code generation tasks. The fundamental bottleneck lies in the crucial speed-quality trade-off. As shown in Figure~\ref {fig:intro}, in code generation tasks, the mainstream DLM sampling strategy can lead to a sharp drop in Pass@1 accuracy (even exceeding 60\%), once it increases parallelism to reduce the sampling steps, making the DLMs nearly unusable. This severe trade-off prevents DLMs from realizing their inherent parallel generation advantages in practice, as the computational savings from fewer steps are offset by a significant drop in quality.

We argue that the root cause of this severe trade-off stems from two fundamental challenges inherent to DLM sampling process: 1) This process exhibits non-uniform difficulty. The complexity of correctly predicting a token varies significantly across the task, generation context, and token position. Therefore, static acceleration strategies per step (such as using a fixed token number or confidence threshold) are suboptimal. They are often overly conservative in simple stages, sacrificing speed, while being overly aggressive in complex stages, significantly degrading quality.
2) This process is particularly susceptible to error propagation. Unlike ARMs, which only decide what the next token is, DLMs must decide both where and what token to generate. An incorrect choice made early in the process, when the contextual information is sparse, becomes permanently "locked in" and cannot be revised. This initial error corrupts the context of all subsequent steps, leading to a cascade of failures from which the model cannot recover.

In this paper, we propose \textbf{Saber}, namely efficient \textbf{S}ampling with \textbf{A}daptive acceleration and \textbf{B}acktracking \textbf{E}nhanced \textbf{R}emasking for DLMs, a novel training-free sampling algorithm designed to address the two fundamental challenges. Specifically, Saber is built on two key strategies: 1) To address non-uniform difficulty, Saber dynamically adjusts the number of tokens generated in parallel at each step, proceeding cautiously in early, context-poor stages and accelerating as more context is established. 2) To counter error accumulation, Saber introduces a backtracking mechanism. It allows the model to reverse tokens that are identified as likely errors based on newly available context, enabling a self-correction process that improves final output quality. 
By introducing these two strategies, Saber achieves substantial speedups while enhancing generation quality.
We further provide theoretical analysis that validates the effectiveness of the proposed method.

To evaluate the effectiveness and generalizability of Saber, we conduct extensive experiments on multiple mainstream code generation benchmarks. We have the findings from the following aspects: 1) Saber achieves the state-of-the-art performance for DLM sampling in code generation, boosting Pass@1 accuracy by an average improvement of 1.9\% over mainstream DLM sampling methods while achieving an average inference speedup of 251.4\%.
2) We demonstrate that Saber is a model-agnostic training-free sampling method, which shows effectiveness on various DLMs under different settings and benchmarks with consistent performance gains. 3) Through a comprehensive ablation study and variants experiments, we validate that both adaptive acceleration and backtracking-enhanced remasking are integral to Saber's success. 4) Furthermore, experiments on mathematical reasoning and scientific reasoning benchmarks demonstrate that Saber generalizes beyond code generation tasks. As a result, Saber effectively mitigates the speed-quality trade-off with lower total computational cost, significantly narrowing the performance gap between DLMs and ARMs for code generation.

\begin{figure*}
    \centering
    \includegraphics[width=\linewidth]{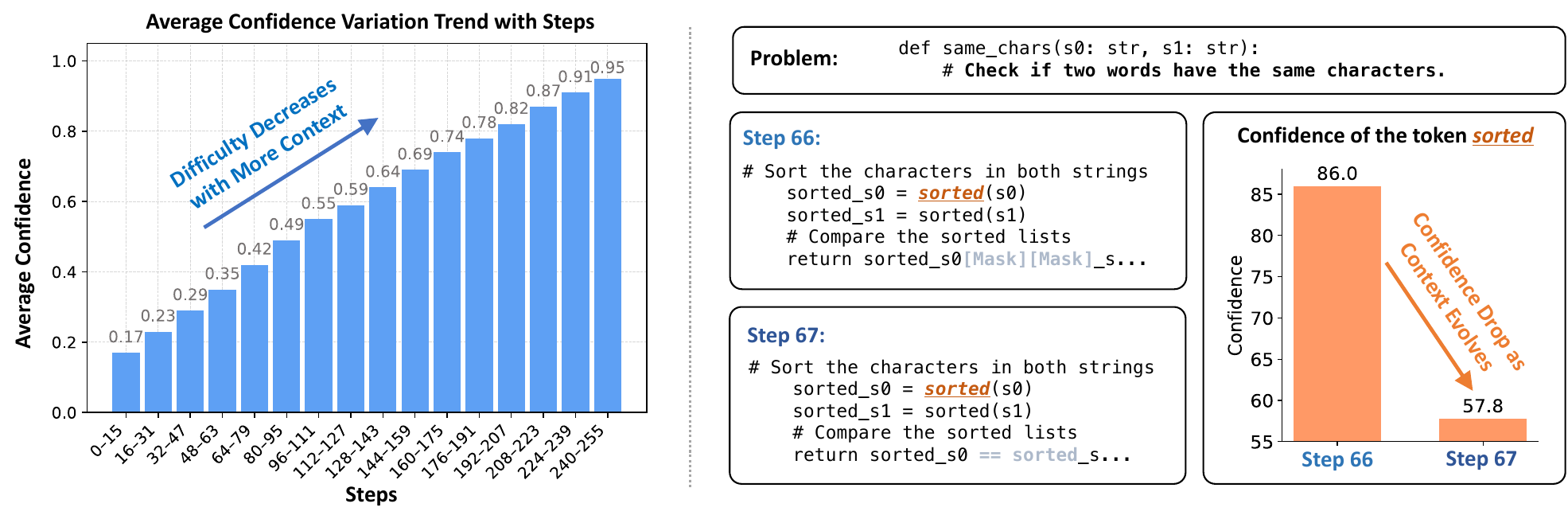}
    \caption{Motivation Example. Left: Average confidence per step. Right: An example of confidence drop for an incorrectly generated token as context evolves during decoding.}
    \label{fig:motivation}
\end{figure*}

\section{Motivation}
\ourapproach~is motivated by two key insights from detailed analyses of the DLM sampling process, as shown in Figure~\ref{fig:motivation}.

\paragraph{Insight 1: Difficulty Decreases over DLM Generation Process.}
The task of generating a masked token is not uniformly difficult throughout the DLM generation process. In the initial steps, the context is sparse, consisting mostly of `\texttt{[MASK]}' tokens and the initial prompt. In this low-information setting, DLMs are highly uncertain, making token generation challenging. However, as more tokens are generated in subsequent steps, the contextual information available to DLMs increases substantially. This richer context progressively reduces DLMs' uncertainty and simplifies the generation of remaining tokens. 

As shown in Figure~\ref{fig:motivation} (left), the DLMs’ average prediction confidence steadily increases as more of the sequence is generated. This observation strongly motivates the need for an adaptive acceleration strategy. An ideal DLM sampler should be cautious when the context is limited and become progressively more aggressive as DLMs' confidence grows. This allows for a more principled approach to acceleration that maximizes speed without prematurely committing to low-confidence tokens.

\paragraph{Insight 2: Dynamic Context of DLM Generated Tokens.}
A significant difference between DLMs and ARMs is the context of generated tokens. In ARMs, the prefix context for each generated token is fixed. However, in DLMs, the context of generated tokens evolves as `\texttt{[MASK]}' tokens are filled in. Therefore, the DLM's predicted confidence of generated tokens can dramatically change as new information becomes available. For example, a token might be predicted with high confidence based on sparse local context, only to be revealed as a likely error once a more complete global context is established, as depicted in Figure~\ref{fig:motivation} (right). 

However, traditional DLM sampling methods are irreversible, i.e., once a token is unmasked, the decision is final and cannot be reversed. This makes them highly susceptible to error propagation, where an overconfident early error corrupts the context for all subsequent steps, leading to a cascade of failures. This issue is a primary driver of the catastrophic collapse when attempting parallel decoding, which highlights the necessity of a backtracking remasking mechanism. By allowing DLMs to revise their own predictions, we can mitigate the risk of early error propagation and enable more robust and aggressive parallel generation. 

\paragraph{Summary.} These two insights reveal a fundamental limitation of current sampling: their static and irreversible design fails to account for the dynamic nature of both generation difficulty and contextual certainty during the DLM sampling process. Therefore, in this paper, we argue that an effective DLM sampler must address these limitations by both adapting its generation speed to the evolving context and being able to revise its own past decisions to mitigate error propagation.

\begin{figure*}
    \centering
    \includegraphics[width=0.95\linewidth]{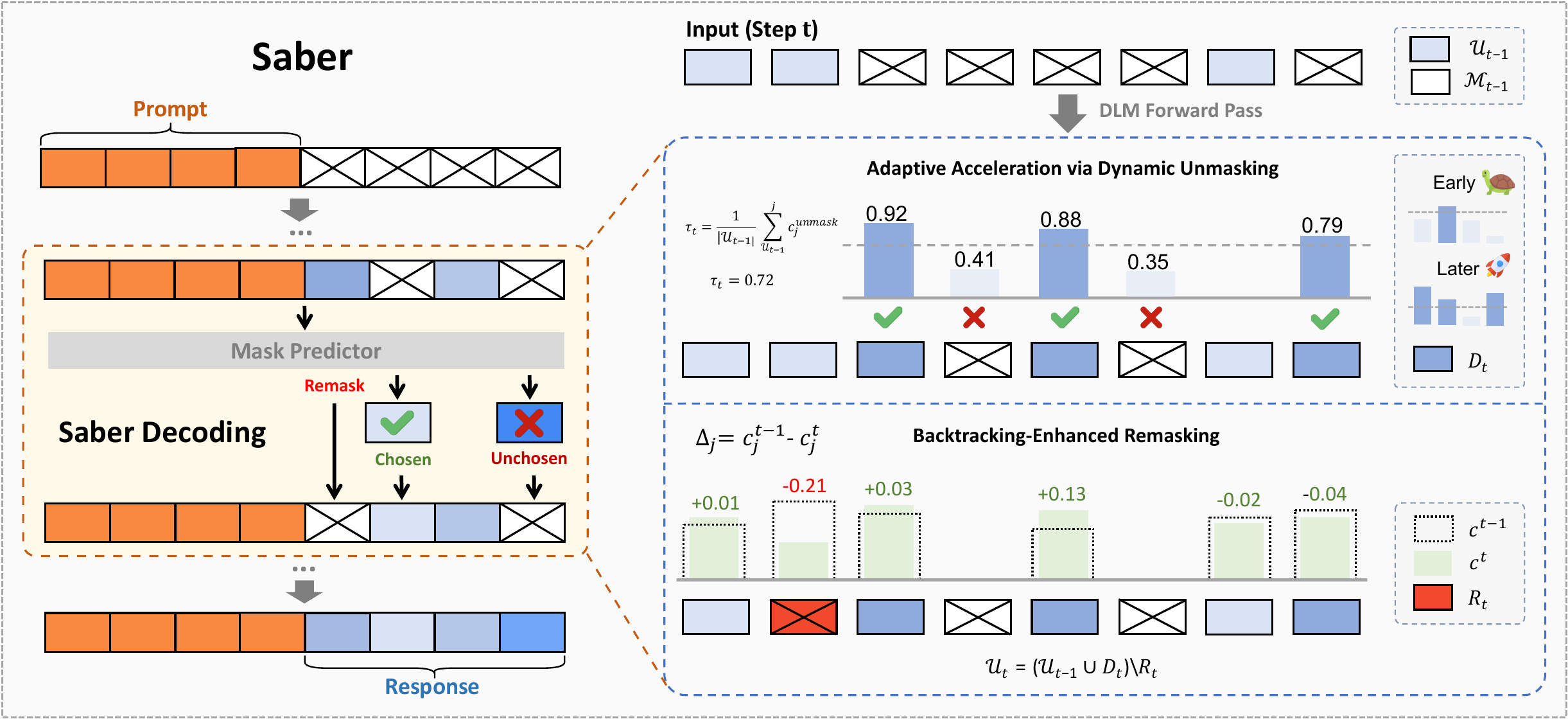}
    \caption{An Overview of Saber in DLM sampling, which consists of two key components, i.e., Adaptive Acceleration via Dynamic Unmasking (AADU) and Backtracking-Enhanced Remasking Mechanism (BERM), during each iterative sampling process.}
    \label{fig:Method}
\end{figure*}

\section{Related Work}
In this section, we outline the two most relevant directions and associated papers of this work.

\subsection{Diffusion Language Models for Code}

The current landscape of language models is dominated by the autoregressive paradigm \cite{Radford2018ImprovingLU, Brown2020LanguageMA, Touvron2023LLaMAOA, Dubey2024TheL3,guo2025deepseek}. However, their strict left-to-right and token-by-token generation process creates a major bottleneck for inference efficiency and inherently limits parallelism \cite{Li2025ASO}. Therefore, a growing body of research on DLMs has emerged \cite{diffusionlm, d3pm, He2022DiffusionBERTIG}, which operate through parallel generation and bidirectional context modeling to address the aforementioned constraints. Recently, large-scale DLMs such as Dream \cite{Ye2025Dream7D}, DiffuLLaMA \cite{diffusionllama}, and LLaDA \cite{llada} have demonstrated performance comparable to similar-scale ARMs, making them a highly promising alternative.

The inherent capabilities of DLMs in global planning and iterative optimization make them naturally suited for code generation \cite{Gong2025DiffuCoderUA, Li2025ASO}. Therefore, the application of DLMs to this domain has become a major research focus \cite{deepmind2025gemini, Gong2025DiffuCoderUA, Xie2025DreamCoder7A, Khanna2025MercuryUL}. However, these works mainly focus on the training process of DLMs, while Saber is a training-free DLM sampling method and is orthogonal to them.

\subsection{Efficient DLM Sampling Methods}
The efficiency of DLMs stems from their ability to generate multiple tokens in parallel \cite{dus, Yu2025DimpleDD,hong2025wide,huang2025pc}. Some studies accelerates this process by setting a fixed threshold, such as Fast-dLLM \cite{Wu2025FastdLLMTA}, WINO \cite{hong2025wide}, and EB-Sampler \cite{ebsampler}. 
However, attempting to unmask multiple tokens in each step degrades the final output quality \cite{Li2025ASO, Zhang2025ASO, Wu2025FastdLLMTA}. Moreover, ReMDM \cite{remdm} proposes a phased sampler that can remask the generated tokens during one of the generation phases. However, the aforementioned methods are less effective on code generation tasks.

To the best of our knowledge, we are the first to combine adaptive acceleration and backtracking enhanced remasking to achieve improvement for both inference speed and output quality in DLM sampling.

\section{Saber}
In this section, we first provide the preliminaries for DLM sampling ($\S~\ref{Preliminaries}$), and then describe the two key components of Saber: Adaptive Acceleration via Dynamic Unmasking ($\S~\ref{AADU}$) and Backtracking-Enhanced Remasking Mechanism ($\S~\ref{BERM}$). 
Finally, we provide the overview of Saber ($\S~\ref{overallSaber}$) in DLM sampling, which is also illustrated in Figure \ref{fig:Method}.

\subsection{Preliminaries}
\label{Preliminaries}

Let a token sequence of length $L$ be denoted by $x = (x_1, \dots, x_L)$, where each token $x_i$ belongs to a vocabulary $\mathcal{V}$. In the diffusion process, we use a special token `\texttt{[MASK]}'. At any denoising step $t$, the sequence $x_t$ consists of a set of unmasked tokens at indices $\mathcal{U}_t$ and a set of masked tokens at indices $\mathcal{M}_t$. The DLM $p_\theta$, parameterized by $\theta$, takes the partially masked sequence $x_t$ as input and outputs a probability distribution over the vocabulary for each masked position $i \in \mathcal{M}_t$. We define the model's confidence in its top prediction for a masked token $i$ as $c_i$:
\begin{equation}
    c_i := \max_{v \in \mathcal{V}} p_\theta(x_i = v \mid x_t), 
\end{equation}
where the mainstream DLM sampling method is to greedily unmask the single token with the highest confidence at each step. Saber improves upon this by the following two key components.

\subsection{Adaptive Acceleration via Dynamic Unmasking}
\label{AADU}

The first component of Saber aims to accelerate inference by unmasking multiple tokens in parallel. Motivated by our observation that the model's prediction difficulty is non-uniform, we introduce a dynamic and adaptive threshold $\tau_t$ to determine which tokens to unmask. This threshold is calculated as the average confidence of all previously unmasked tokens:
\begin{equation}
\tau_t = 
\begin{cases} 
\frac{1}{|\mathcal{U}_{t-1}|} \sum_{j \in \mathcal{U}_{t-1}} c_j^{\text{unmask}} & \text{if } t>0, \\
c_{max} & \text{otherwise,}
\end{cases}
\end{equation}
where $c_j^{\text{unmask}}$ is the confidence score of token $j$ at the step it was unmasked, and we initialize $\tau_0$ to $c_{max}$ for the initial step.

This dynamic threshold naturally encourages a cautious-to-aggressive decoding trajectory. In early steps, when the context is sparse and average confidence is low, $\tau_t$ is low, allowing only the most certain tokens to be unmasked. As more high-confidence tokens are generated, $\tau_t$ rises, permitting more aggressive parallel unmasking in later, more context-rich stages. Using the threshold $\tau_t$, we identify a set of candidate tokens to be drafted, $\mathcal{D}_t$, which includes all masked tokens whose current confidence exceeds $\tau_t$:
\begin{equation}
    \mathcal{D}_t = \{ i \in \mathcal{M}_{t-1} \mid c_i \geq \tau_t \},
\end{equation}
where the tokens are provisionally unmasked with their most likely prediction.

\begin{algorithm}[ht!]
\caption{Pseudocode of Saber in each step.}
\label{alg:saber_step}
\begin{algorithmic}[1]
\State \textbf{Input:} Sequence $x_{t-1}$, DLM $p_\theta$, unmasked indices $\mathcal{U}_{t-1}$, unmasked confidences $c^{\text{unmask}}$
\State \textbf{Output:} Updated sequence $x_t$
\Statex // \textbf{S1: Adaptive Acceleration}
\State Compute confidences $c_i$ for all $i \in \mathcal{M}_{t-1}$ using $p_\theta(\cdot \mid x_{t-1})$.
\If{$t > 0$}
    \State $\tau_t \leftarrow \frac{1}{|\mathcal{U}_{t-1}|} \sum_{j \in \mathcal{U}_{t-1}} c_j^{\text{unmask}}$.
\Else
    \State $\tau_t \leftarrow c_{max}$.
\EndIf
\State $\mathcal{D}_t \leftarrow \{i \in \mathcal{M}_{t-1} \mid c_i > \tau_t\}$.
\State Create candidate sequence $x_t'$ by unmasking tokens in $\mathcal{D}_t$.
\Statex // \textbf{S2: Backtracking-Enhanced Remasking}
\State $\mu_t \leftarrow \max(1, \lfloor |\mathcal{D}_t| / \mu \rfloor)$.
\State Re-evaluate confidences $c_j^t$ for all $j \in \mathcal{U}_{t-1}$ using the new context $p_\theta(\cdot \mid x_t')$.
\State Initialize an empty set for confidence drops $\Delta$.
\For{each token $j \in \mathcal{U}_{t-1}$}
    \State $\Delta_j \leftarrow c_j^{t-1} - c_j^t$. \Comment{Calculate drop}
    \State Add $(j, \Delta_j)$ to $\Delta$.
\EndFor
\State $\mathcal{R}_t \leftarrow$ indices of the $\mu_t$ tokens from $\Delta$ with the largest drop.
\State Create final sequence $x_t$ by re-masking tokens at indices $\mathcal{R}_t$ in $x_t'$.
\State Update $c^{\text{unmask}}$ by removing confidences for $j \in \mathcal{R}_t$ and adding confidences for $i \in \mathcal{D}_t$.
\State $\mathcal{U}_t \leftarrow (\mathcal{U}_{t-1} \cup \mathcal{D}_t) \setminus \mathcal{R}_t$.
\State \textbf{return} $x_t$, $\mathcal{U}_{t}$, $c^{\text{unmask}}$.
\end{algorithmic}
\end{algorithm}

\subsection{Backtracking-Enhanced Remasking Mechanism}
\label{BERM}

The second component of Saber introduces a backtracking mechanism to correct for potential errors made during the aggressive generation in the previous stage. This step is crucial for preventing the error propagation that causes performance collapse.

Unlike methods that use a fixed threshold, Saber's backtracking mechanism first determines the number of tokens to revise, $\mu_t$, based on how aggressively it generated tokens in the current step:
\begin{equation}
    \mu_t = \max(1, \lfloor |\mathcal{D}_t| / \mu \rfloor),
\end{equation}
where $|\mathcal{D}_t|$ is the size of the newly unmasked set and $\mu$ is a hyperparameter. This ensures that we revise at least one token while limiting the revision to a small fraction of the current step's output to maintain speed.

Then, we identify which tokens to revise by focusing on those previously unmasked tokens that are most inconsistent with the newly available context. For each existing token $j \in \gU_{t-1}$, we compute its confidence drop, $\Delta_j$, defined as the difference between the unmasked confidences of (t-1)-th time $c_j^{t-1}$, and its re-evaluated confidence at the current step $c_j^t$:
\begin{equation}
    \Delta_j = c_j^{t-1} - c_j^t,
\end{equation}
where a large $\Delta_j$ indicates that the model's confidence in its earlier prediction has significantly weakened. We then identify the set of tokens to be reversed, $\gR_t \subseteq \gU_{t-1}$, by selecting the $\mu_t$ tokens that exhibit the largest confidence drop. These are the tokens the model has the most regret about, and they are reverted to `\texttt{[MASK]}' to be reconsidered in future steps with a richer context.

\subsection{Overall Procedure of Saber}
\label{overallSaber}
At the conclusion of each step $t$, the final set of unmasked tokens is updated by integrating the outcomes of both the adaptive acceleration and backtracking stages:
\begin{equation}
    \mathcal{U}_t = (\mathcal{U}_{t-1} \cup \mathcal{D}_t) \setminus \mathcal{R}_t.
\end{equation}

By combining adaptive acceleration with an efficient backtracking mechanism, Saber can decode aggressively while pruning the most probable errors, thus achieving a superior balance between inference speed and generation quality. The pseudocode of Saber in each DLM sampling step is summarized in Algorithm~\ref{alg:saber_step}.

\section{Theoretical Analysis}
\label{sec:theoretical_analysis}

To formalize the intuition behind the Saber algorithm and explain why it mitigates the failure of standard DLMs in code generation, we analyze the DLM decoding process by modeling the evolution of errors across sampling steps.

\paragraph{Error Accumulation in the Standard DLMs.}
Given the strict structural dependencies of programming languages, early generation errors severely contaminate the bidirectional context, irreversibly amplifying the model's prediction uncertainty for subsequent tokens. Let $e_t$ be the random variable representing the number of structural errors rigidly locked in the unmasked context at step $t$, and let $\err(\mathcal{D}_t) = \sum_{k \in \mathcal{D}_t} \mathbb{I}(\err_k)$ denote the number of errors within the newly unmasked candidate set $\mathcal{D}_t$. In traditional static parallel DLM sampling, the error evolution is purely monotonic:
\begin{equation}
\label{eq:standard_error}
\E[e_t \mid x_{t-1}] = \E[e_{t-1} \mid x_{t-1}] + \E[\err(\mathcal{D}_t) \mid x_{t-1}].
\end{equation}
Saber addresses this through two mechanisms: Adaptive Acceleration via Dynamic Unmasking (AADU) and the Backtracking-Enhanced Remasking Mechanism (BERM).

\paragraph{AADU (Bounding New Errors).}
By utilizing a dynamic threshold $\tau_t$ to conservatively constrain the unmasking candidate set $\mathcal{D}_t$, AADU establishes a dynamic fault-tolerance boundary. Assuming the DLM's subjective confidence serves as a proxy for token accuracy with an allowable calibration slack (formalized in Appendix \ref{app:assumptions}), we establish the following bound.

\begin{lemma}[Step-wise Error Injection Bounding]
\label{lem:aadu}
Given the context $x_{t-1}$ and a model overconfidence slack $\delta_{\text{calib}} \ge 0$, the conditional expected number of newly injected errors is bounded by:
\begin{equation}
\E[\err(\mathcal{D}_t) \mid x_{t-1}] \le |\mathcal{D}_t|(1 - \tau_t + \delta_{\text{calib}}).
\end{equation}
\end{lemma}
While Lemma \ref{lem:aadu} holds for any threshold, our specific choice of using the historical average of unmasked tokens for $\tau_t$ serves as an effective empirical way to track the model's evolving certainty, safely expanding $|\mathcal{D}_t|$ only when the context becomes reliable.

\paragraph{BERM (Pruning Contextual Conflicts).}
While AADU bounds the injection of \textit{new} errors, BERM is designed to mitigate the strict monotonic accumulation of \textit{past} errors, as shown in Eq. (\ref{eq:standard_error}). It evaluates the step-wise confidence drop (i.e., $\Delta_j$) of previously generated tokens to identify contextual conflicts (see Appendix \ref{app:assumptions}).

\begin{proposition}[Degradation Pruning]
\label{prop:berm}
By selecting the $\mu_t$ tokens with the largest confidence drop to form the rollback set $\mathcal{R}_t$, BERM selects the subset of historical context that minimizes the upper bound of the total step-wise reliability degradation within the retained context for a given pruning budget $\mu_t$.
\end{proposition}

\paragraph{System Evolution under Saber.}
By combining AADU and BERM, Saber aims to transform the monotonically increasing error accumulation into a more controlled process. Parallel unmasking inherently introduces a contextual collision penalty bounded by $\epsilon$ (Assumption \ref{ass:collision}), stemming from ignored conditional dependencies. Because $\epsilon$ scales quadratically with the unmasked set size $|\mathcal{D}_t|$, it could easily overwhelm a linear pruning budget. Here, the synergy of Saber is critical: AADU bounds $|\mathcal{D}_t|$ to prevent $\epsilon$ from exploding, while BERM actively offsets the remaining structural errors. Relying on assumptions regarding local reliability preservation (Assumption \ref{ass:non-degradation}), as derived in Appendix \ref{app:evolution}, by dynamically bounding the new errors (via AADU) and actively pruning the most conflicting tokens (via BERM), the evolution of Saber's expected error bound can be formulated as:
\begin{align}
&\mathbb{E}[e_t^{\text{Saber}} \mid x_t] \le \mathbb{E}[e_{t-1} \mid x_{t-1}] \nonumber \\
& - \underbrace{\sum_{i \in \mathcal{R}_t}\mathbb{P}(\err_i \mid x_{t-1})}_{\text{BERM Pruning}} + \underbrace{\sum_{k \in \mathcal{D}_t}\mathbb{P}(\err_k \mid x_{t-1})}_{\text{AADU Bounding}} + \epsilon. \nonumber
\end{align}

\paragraph{Superiority over Traditional Bounds.}
In traditional static parallel DLM sampling, the expected error accumulates monotonically without any pruning mechanism. In contrast, Saber disrupts this monotonic accumulation, which can achieve a lower theoretical error bound, $\mathbb{E}[e_t^{\text{Saber}} \mid x_t] < \mathbb{E}[e_t^{\text{Traditional}} \mid x_t]$. We empirically observe and validate this penalty-offsetting condition in our ablation studies (Section 6). The detailed mathematical formulation of the traditional bound and the comparative proof are provided in Appendix \ref{app:superiority}.

\paragraph{Summary.}
In strong structural constraints tasks, such as code generation, the underlying structural dependencies are highly non-linear. A single early token error catastrophically cascades, causing the entire structure to collapse. Therefore, keeping the sequence entirely error-free ($e_t = 0$) throughout decoding is paramount. By tightly bounding early error injection and actively pruning degraded tokens to offset parallel collision penalties, Saber effectively preserves the code's structural integrity. 
We note that this analytical framework operates at the per-step level. The end-to-end error dynamics throughout the generation trajectory are validated by our subsequent experiments.

\begin{table*}[t]
\caption{Comparison of Saber and existing DLM sampling methods, where the \textbf{bold} indicates the best performance in this column while the \underline{underline} indicates the second-best performance, and ET means the Pass@1 performance on its extended test case version.}
\label{tab:main_results}
\resizebox{\textwidth}{!}{
\begin{tabular}{@{}lrrrrrlrrrrr@{}}
\toprule
\multirow{2}{*}{Method} & \multicolumn{4}{c}{HumanEval}   & \multicolumn{4}{c}{MBPP}             & \multicolumn{3}{c}{LiveCodeBench}  \\ \cmidrule(l){2-5} \cmidrule(l){6-9} \cmidrule(l){10-12}
& Pass@1 $\uparrow$ & ET $\uparrow$ & Step $\downarrow$ & Time $\downarrow$ & Pass@1 $\uparrow$ & ET $\uparrow$ & Step $\downarrow$ & Time $\downarrow$ & Pass@1 $\uparrow$ & Step $\downarrow$ & Time $\downarrow$ \\ \midrule
\multicolumn{12}{l}{\textbf{{{\textcolor{gray}{Standard DLM Sampling}}}}}                            \\
Random                  & 14.63                     & 12.80                  & 256                      & 1:29:40                  & 22.95                     & 18.26                 & 256                      & 2:51:28                  & 0      & 256                      & 4:09:49                  \\
Entropy                 & 41.46                     & 34.15                 & 256                      & 1:30:22                  & 42.15                     & 31.14                 & 256                      & 2:56:42                  & 4.00                       & 256                      & 4:30:31                  \\
Confidence              & \underline{43.29}                     & \underline{35.79}           & 256                      & 2:11:52                  & 42.86                     & 31.38                 & 256                      & 3:12:08                  & \underline{9.75}                     & 256                      & 5:59:07                  \\
\hdashline
\multicolumn{12}{l}{\textbf{{{\textcolor{gray}{Efficient DLM Sampling}}}}}                                    \\
Confidence (p=2)      & 34.76                     & 28.66                 & 128                      & 51:13                 & 40.75                     & 28.57                 & 128                      & 1:35:13                  & 9.25                     & 128                      & 2:57:16                  \\
SAR   (p=2)             & 35.98                     & 29.27                 & 128                      & 1:33:00            & 40.05                     & 27.86                 & 128                      & 1:36:05                  & 9.50                      & 128                      & 2:57:17                  \\
Fast-dLLM        & 39.63                     & 34.15                 & 256                      & 59:40                 & \underline{44.03}               & 30.44                 & 256                      & 2:30:24                  & 8.75                     & 256                      & 2:33:29            \\
Fast-dLLM (+parallel)       & 39.63                     & 33.54                 & \textbf{96.24}                    & \textbf{25:25}                    & 39.34               & 27.63                 & \textbf{73.13}                    & \textbf{43:18} & 2.30                      & \underline{96.28}                    & \textbf{43:22} \\        
ReMDM                   & 20.73                     & 18.29                  & 128                      & 1:26:50                  & 31.62                     & 22.48                 & 128                      & \underline{1:28:51}         & 3.30                          & 128                        & 2:50:23                        \\
WINO                    & 40.24               & 31.71                 & \underline{100.12}          & 57:10               & 43.09                     & \underline{31.38}           & \underline{88.49}           & 1:44:51                  & 9.25                     & \textbf{77.43}           & 2:40:30                  \\
\hdashline
Saber                   & \textbf{45.12}            & \textbf{35.98}        & 118.92             & \textbf{41:55}        & \textbf{44.73}            & \textbf{33.02}        & 110.96             & 1:33:33            & \textbf{11.00}              & 122.47             & \underline{2:33:17}         \\ \bottomrule
\end{tabular}}
\end{table*}

\section{Experimental Results}
\label{sec:experiments}

In this section, we present a comprehensive empirical evaluation of \ourapproach. We first compare its performance and efficiency against a wide range of existing DLM sampling methods on multiple code generation benchmarks (\S\ref{sec:main_results}). Next, we demonstrate the model-agnostic nature of \ourapproach~by applying it to various state-of-the-art DLMs (\S\ref{sec:generalizability}). Finally, we conduct a detailed ablation study to dissect the individual contributions of our proposed components (\S\ref{sec:ablation}) and provide the discussion of Saber (\S\ref{sec:qualitative}). The detailed description of experiment setups can be found in Appendix \ref{app_setup}.

\subsection{Main Results}
\label{sec:main_results}

Table~\ref{tab:main_results} presents the main results of our comparison on the HumanEval, MBPP, HumanEval-ET and MBPP-ET, and LiveCodeBench datasets. The findings clearly demonstrate that \ourapproach~sets a new state-of-the-art for DLM sampling in code generation, achieving the highest Pass@1 scores across all benchmarks while simultaneously delivering substantial improvements in inference speed.

\textbf{\ourapproach~Effectively Mitigates the Speed-Quality Trade-off.} Compared to standard DLM sampling strategies (Random, Entropy, Confidence), \ourapproach~delivers vastly superior performance. For instance, on HumanEval, \ourapproach~improves the Pass@1 score from 43.3\% (Confidence) to 45.1\% while reducing the inference time by nearly 70\% (from over 2 hours to just 41 minutes). This result directly refutes the notion that acceleration must come at the cost of quality. While naively increasing parallelism by generating more tokens per step (e.g., Confidence p=2) leads to a significant performance drop (from 43.3\% to 34.8\%), \ourapproach's intelligent sampling process successfully avoids this collapse.

\textbf{\ourapproach~Outperforms State-of-the-Art Efficient Samplers.} When compared to recent efficient sampling methods, \ourapproach~establishes a new Pareto frontier for the speed-quality trade-off. WINO, a strong baseline, achieves impressive speed by minimizing decoding steps. However, \ourapproach~is even faster in terms of time on most benchmarks, indicating a more efficient computation per step. For example, on HumanEval, \ourapproach~is over 25\% faster than WINO while also achieving a $\sim$5\% higher Pass@1 score. This superior performance is attributed to our backtracking mechanism, which provides a safety net for the adaptive acceleration, allowing for aggressive parallelization without sacrificing accuracy. Similarly, while Fast-dLLM shows competitive results on MBPP, \ourapproach~matches its quality while being nearly 40\% faster. On LiveCodeBench, a benchmark designed to be robust against contamination, \ourapproach~also achieves the state-of-the-art performance, demonstrating its strong generalization capabilities.

Overall, these results confirm that \ourapproach~successfully breaks the existing speed-quality compromise in DLM sampling for code generation.

\subsection{Generalizability Across Different DLMs}
\label{sec:generalizability}

To validate the model-agnostic claim of \ourapproach, we apply it to three distinct open-source DLMs, i.e., \texttt{LLaDA-8B-Instruct} \cite{llada}, \texttt{Dream-v0-Instruct-7B} \cite{Ye2025Dream7D}, and \texttt{DiffuCoder-7B-cpGRPO} \cite{Gong2025DiffuCoderUA}. We compare the performance of Saber against the standard confidence-based sampler for each DLM on the HumanEval benchmark.

\begin{table}[h]
\caption{Effectiveness of Saber compared to mainstream DLM sampling method based on different DLMs.}
\label{tab:generalizability}
\centering
\small
{
\begin{tabular}{@{}lrrr@{}}
\toprule
              & \multicolumn{1}{c}{Pass@1 $\uparrow$} & \multicolumn{1}{c}{Steps $\downarrow$} & \multicolumn{1}{c}{Time $\downarrow$}  \\ \midrule
LLaDA-8B-Instruct    & \multicolumn{1}{l}{}       & \multicolumn{1}{l}{}     & \multicolumn{1}{l}{}      \\
Confidence (p=1)       & 0.4329                                & 256         & 2:11:52                     \\
Saber         & \textbf{0.4512}                                 & \textbf{118.92}         & \textbf{41:55}                \\
\hdashline
Dream-v0-Instruct-7B   & \multicolumn{1}{l}{}       & \multicolumn{1}{l}{}     & \multicolumn{1}{l}{}      \\
Confidence (p=1)       & 0.2805                        & 256       & 1:16:15                               \\
Saber         & \textbf{0.2927}                                & \textbf{156.68}      & \textbf{46:39}                    \\
\hdashline
DiffuCoder-7B-cpGRPO	 & \multicolumn{1}{l}{}       & \multicolumn{1}{l}{}     & \multicolumn{1}{l}{}      \\
Confidence (p=1)      & 0.5671                                     & 256          & 1:12:47               \\
Saber         & \textbf{0.5732}                                      & \textbf{140.34}        & \textbf{37:08}            \\ \bottomrule
\end{tabular}}
\end{table}

As shown in Table~\ref{tab:generalizability}, \ourapproach~consistently improves both accuracy and efficiency across all tested models, demonstrating that its benefits are not tied to a specific architecture or training process. For each model, \ourapproach~delivers a higher Pass@1 score while simultaneously reducing the number of decoding steps and the total inference time. For instance, on \texttt{Dream-v0-Instruct-7B}, \ourapproach~boosts Pass@1 and cuts inference time by nearly 40\%. On \texttt{DiffuCoder-7B}, a model specifically optimized for code, \ourapproach~further enhances its performance while reducing the inference time by nearly half.

To evaluate the generalizability of Saber from more dimensions, we conduct experiments on different settings, domains, and benchmarks in Appendix \ref{DifferentConfigurations}, Appendix \ref{DifferentDomain}, and Appendix \ref{differentbenchmark}, respectively. Notably, beyond code generation, Saber also demonstrates consistent improvements on mathematical reasoning and scientific reasoning tasks (See Appendix \ref{DifferentDomain}). This robust performance across different model families, settings, domains, and benchmarks validates that \ourapproach~addresses fundamental challenges in DLM sampling, making it a general, plug-and-play enhancement.

\subsection{Ablation Study}
\label{sec:ablation}

To understand the individual contributions of the two core components of \ourapproach, i.e., Adaptive Acceleration via Dynamic Unmasking and Backtracking-Enhanced Remasking Mechanism, we conduct a thorough ablation study on the HumanEval dataset. The results are presented in Table~\ref{tab:ablation}.

\begin{table}[h]
\caption{Ablation study of different components in our proposed method.}
\label{tab:ablation}
\centering
\small
{
\begin{tabular}{@{}lrrr@{}}
\toprule
Method & \multicolumn{1}{c}{Pass@1 $\uparrow$} & \multicolumn{1}{c}{Steps $\downarrow$} & \multicolumn{1}{c}{Time $\downarrow$} \\ \midrule
Ours & \textbf{0.4512} & \underline{118.92} & \underline{41:55} \\
w/o Adaptive Accelerate & \underline{0.4451}& 256 & 1:32:33 \\
w/o Backtracking Remask & 0.3523 & \textbf{65.67} & \textbf{28:30}\\
w/o both & 0.3476 & 128 & 51:13 \\
\hdashline
$\Delta$ confidence from init. & 0.4207 & 121.46 & 42:32 \\
 \bottomrule
\end{tabular}
}
\end{table}

\textbf{Adaptive Acceleration is the Primary Driver of Efficiency.} When we remove the Adaptive Acceleration via Dynamic Unmasking, the sampler relies solely on the backtracking mechanism. While the Pass@1 score remains high at 44.5\%, the number of decoding steps reverts to the baseline 256, and the inference time increases dramatically to over 90 minutes. This clearly demonstrates that the adaptive acceleration component is the main source of \ourapproach's speedup.

\textbf{Backtracking is Essential for High Quality.} Conversely, when we remove Backtracking-Enhanced Remasking Mechanism, the sampler becomes a purely aggressive adaptive accelerator. This variant is extremely fast, finishing in under 30 minutes with only 65.67 steps on average. However, this speed comes at a steep price: the Pass@1 score drops significantly from 45.1\% to 35.23\%. This result highlights that aggressive parallelization without a corrective mechanism is prone to error propagation, confirming that the backtracking stage is crucial for maintaining high generation quality.

\textbf{Synergy of Components.} \ourapproach~achieves the best of both worlds, i.e., a high Pass@1 score of 45.1\% and a fast inference time of $\sim$41 minutes, which also shows that the two components are synergistic. The adaptive acceleration allows for aggressive sampling, while the backtracking mechanism provides the necessary safety net to prune errors, enabling a combination of speed and accuracy that neither component can achieve alone. We also validate our dynamic thresholding strategy by replacing it with the average threshold of init generation of tokens ($\Delta$ confidence from init.). This results in a lower Pass@1 score of 42.1\%, confirming the benefits of an adaptive approach that adjusts to the evolving context.

\subsection{Qualitative Analysis}
\label{sec:qualitative}
\paragraph{Error Type Analysis.} 
We conduct an error type analysis and categorize the generation failures into three distinct types: Syntax Errors, Compilation/Runtime Errors, and Semantic Errors. As shown in Table~\ref{tab:error_type}, Saber consistently outperforms the standard baseline across all error categories. Notably, Saber significantly reduces Syntax Errors by $66.7\%$ and Compilation/Runtime Errors by $21.5\%$. This quantitative evidence robustly supports our theoretical assertion that the backtracking-enhanced remasking mechanism effectively mitigates structural and syntactical error propagation during parallel decoding.

\begin{table}[htbp]
\centering
\small
\caption{Error type analysis.}
\label{tab:error_type}
\begin{tabular}{lccc}
\toprule
Error Type & Baseline & Saber & $\downarrow\Delta$ \\
\midrule
Syntax Error              & 3  & \textbf{1}  & 66.7\% \\
Compilation/Runtime Error & 14 & \textbf{11} & 21.5\% \\
Semantic Error            & 92 & \textbf{73} & 20.7\% \\
\bottomrule
\end{tabular}
\end{table}

\paragraph{Case Study.} Figure~\ref{fig:qualitative} presents a side-by-side comparison of code generated by the default LLaDA sampler and \ourapproach~on two problems from the HumanEval benchmark. These examples highlight how \ourapproach's ability to self-correct prevents the kind of logical failures that plague standard irreversible samplers. 

In Problem 1, the default sampler produces code, which is syntactically plausible but logically nonsensical. In contrast, \ourapproach~generates the correct, standard nested loop structure. This suggests that the iterative refinement process, guided by backtracking, helps enforce logical and structural coherence, which is paramount in code generation. In Problem 2, the default sampler fundamentally misunderstands the problem's constraints. \ourapproach, however, correctly decomposes the problem into its core logical components: checking the array's length and verifying the occurrence count of the maximum element. This ability to correctly construct multi-step, constraint-based logic is a direct benefit of the backtracking mechanism. We hypothesize that the model may initially draft a simpler, incorrect solution, which is then revised in subsequent steps as the evolving context makes the error more apparent, leading to the robust final code.

\section{Conclusion}
In this paper, we addressed the critical speed-quality trade-off for DLM sampling in code generation and introduced Saber, a novel, training-free sampling algorithm for DLMs that combines both adaptive acceleration via dynamic unmasking and backtracking-enhanced remasking mechanism. Our extensive experiments indicate that Saber substantially outperforms existing DLM sampling methods with great generalizability, significantly narrowing the performance gap with autoregressive models in code generation. Moreover, results on mathematical reasoning and scientific reasoning benchmarks suggest that Saber generalizes beyond code generation. We leave a broader exploration across diverse domains and tasks as future work.

\bibliography{ref}
\bibliographystyle{preprint}

\newpage
\appendix
\onecolumn
\section{Extended Analytical Formulation and Proofs}

\subsection{Formal Assumptions and Definitions}
\label{app:assumptions}

\textit{Throughout our analysis, $\mathbb{P}(\text{err}_i \mid x)$ denotes the \textbf{true objective error probability} at position $i$ under the reference ground truth distribution.}

\begin{assumption}[Tight Confidence Calibration with Slack]
\label{ass:calibration}
We assume the DLM's predicted confidence score provides a tight two-sided bound on the true error probability evaluated under any arbitrary conditioning context $x$, subject to a calibration slack $\delta_{\text{calib}} \ge 0$. Specifically, the conditional probability of a prediction error at position $i$ satisfies:
\begin{equation}
|\Pbb(\err_i \mid x) - (1 - c_i^x)| \le \delta_{\text{calib}},
\end{equation}
where $c_i^x$ is the predicted confidence given context $x$, i.e., $p_\theta(x_i = y_i^* \mid x)$. (for simplicity in our formulations, we use $c_i^t$ to denote the confidence given context $x_t$, and $c_i^{t-1}$ given $x_{t-1}$).

\textit{While modern neural networks consistently exhibit context- and position-dependent calibration issues, explicitly bounding the deviation globally via $\delta_{\text{calib}}$ serves as a simplifying mathematical assumption. In practice, assuming equality holds in expectation, variations in $\delta_{\text{calib}}$ are implicitly absorbed by our conservative dynamic thresholding strategy.}
\end{assumption}

\begin{definition}[Step-wise Reliability Degradation]
For a historical token $j \in \mathcal{U}_{t-1}$, the degradation in reliability between the previous step $t-1$ and the current step $t$ is quantified by its step-wise confidence drop, defined as:
\begin{equation}
\Delta_j = c_j^{t-1} - c_j^t.
\end{equation}
A larger $\Delta_j$ indicates that the model's certainty regarding the previously generated token $j$ has been significantly weakened by the newly established context $x_t$.
\end{definition}

\begin{assumption}[Local Reliability Preservation via Conflict Resolution]
\label{ass:non-degradation}
A high confidence drop ($\Delta_j$) indicates that a previously generated token $j$ is highly incompatible with the newly decoded context $\mathcal{D}_t$. We assume that by removing the $\mu_t$ most conflicting tokens ($\mathcal{R}_t$), BERM acts as a localized coordinate descent step. Driven by the Maximum A Posteriori (MAP) objective of the masked diffusion model, we assume this targeted pruning resolves the largest contextual contradictions, ensuring the expected reliability of the remaining retained context ($\mathcal{U}_{t-1} \setminus \mathcal{R}_t$) does not severely deteriorate:
\begin{equation}
\sum_{j \in \mathcal{U}_{t-1} \setminus \mathcal{R}_t} \Pbb(\err_j \mid x_t) \le \sum_{j \in \mathcal{U}_{t-1} \setminus \mathcal{R}_t} \Pbb(\err_j \mid x_{t-1}).
\end{equation}
\textit{While masking tokens can theoretically disrupt long-range syntactic dependencies in non-autoregressive generation, we treat this targeted pruning as a first-order approximation that primarily resolves severe local structural conflicts without catastrophically destabilizing the broader context.}
\end{assumption}

\begin{assumption}[Bounded Contextual Collision in Parallel Unmasking]
\label{ass:collision}
Unlike purely autoregressive decoding, unmasking multiple tokens ($\mathcal{D}_t$) simultaneously introduces joint dependencies. We encapsulate this structural collision penalty as $\epsilon$, which is upper-bounded by a quadratic function of the unmasked set size:
\begin{equation}
\epsilon \le C \cdot |\mathcal{D}_t|^2,
\end{equation}
where $C > 0$ is a constant representing the maximum pairwise mutual information penalty. In fully autoregressive decoding, the joint probability incorporates full dependencies: $p(x_A, x_B) = p(x_A)p(x_B \mid x_A)$. Parallel unmasking simplifies this by assuming local conditional independence: $p(x_A, x_B) \approx p(x_A)p(x_B)$. Since a set of size $|\mathcal{D}_t|$ contains $\binom{|\mathcal{D}_t|}{2}$ interacting pairs, the structural collision approximation error naturally scales quadratically. \textit{This highlights the necessity of AADU: without dynamically bounding $|\mathcal{D}_t|$, the quadratic growth of $\epsilon$ would inevitably exceed the BERM's linear pruning capacity.}
\end{assumption}

\subsection{Detailed System Evolution Derivation}
\label{app:evolution}
In sequence generation, the error probability of any token is strongly conditioned on the current context. When transitioning from step $t-1$ to $t$, the context evolves from $x_{t-1}$ to $x_t$ based on the updated unmasked set $\mathcal{U}_t = (\mathcal{U}_{t-1} \setminus \mathcal{R}_t) \cup \mathcal{D}_t$. Let $e_t^{\text{Saber}} = \sum_{m \in \mathcal{U}_t} \mathbb{I}(\err_m)$ denote the total number of errors in the unmasked set at step $t$.

By taking the conditional expectation with respect to the \textit{updated} context $x_t$, applying the contextual preservation (Assumption \ref{ass:non-degradation}), and incorporating the parallel collision penalty (Assumption \ref{ass:collision}), the upper bound of the system error at the end of step $t$ is derived as:
\begin{align}
\E[e_t^{\text{Saber}} \mid x_t] &= \sum_{j \in \mathcal{U}_{t-1} \setminus \mathcal{R}_t} \Pbb(\err_j \mid x_t) + \sum_{k \in \mathcal{D}_t} \Pbb(\err_k \mid x_t) \nonumber \\
&\le \sum_{j \in \mathcal{U}_{t-1} \setminus \mathcal{R}_t} \Pbb(\err_j \mid x_{t-1}) + \sum_{k \in \mathcal{D}_t} \Pbb(\err_k \mid x_{t-1}) + \epsilon \nonumber \\
&= \E[e_{t-1} \mid x_{t-1}] - \sum_{i \in \mathcal{R}_t} \Pbb(\err_i \mid x_{t-1}) + \sum_{k \in \mathcal{D}_t} \Pbb(\err_k \mid x_{t-1}) + \epsilon.
\end{align}
Notice that the transition to the final equality holds by the linearity of expectation, where the total expected error at $t-1$ encapsulates the sum of probabilities. The third term $\sum_{k \in \mathcal{D}_t} \Pbb(\err_k \mid x_{t-1})$ is bounded by $|\mathcal{D}_t|(1 - \tau_t + \delta_{\text{calib}})$ per Lemma \ref{lem:aadu}, while the pruning term is designed to offset the $\epsilon$ collision penalty.

\subsection{Comparison with Traditional Bounds}
\label{app:superiority}
In traditional static parallel DLM sampling, the expected error accumulates monotonically without any pruning mechanism. Given a candidate set $\mathcal{D}_t$, the error bound at step $t$ is:
\begin{equation}
\mathbb{E}[e_t^{\text{Traditional}} \mid x_t] \le \mathbb{E}[e_{t-1} \mid x_{t-1}] + |\mathcal{D}_t|(1 - \tau_{\text{static}} + \delta_{\text{calib}}) + \epsilon,
\end{equation}
where $\epsilon$ is the corresponding collision penalty, and $\tau_{\text{static}}$ denotes the fixed confidence threshold used by standard parallel decoding baselines.

In contrast, Saber disrupts this monotonicity. By dynamically bounding the new errors (via AADU) and actively pruning the most conflicting tokens (via BERM), Saber's expected error satisfies:
\begin{align}
\mathbb{E}[e_t^{\text{Saber}} \mid x_t] \le \mathbb{E}[e_{t-1} \mid x_{t-1}] - \sum_{i \in \mathcal{R}_t}\mathbb{P}(\err_i \mid x_{t-1}) + \sum_{k \in \mathcal{D}_t}\mathbb{P}(\err_k \mid x_{t-1}) + \epsilon.
\end{align}
Because the pruning term explicitly offsets the collision penalty $\epsilon$, and the dynamic thresholding restricts the injection of new errors, under the condition that $\sum_{i \in \mathcal{R}_t}\mathbb{P}(\err_i \mid x_{t-1}) > |\mathcal{D}_t|(\tau_{\text{static}} - \tau_t)$, Saber can achieve a lower error bound $\mathbb{E}[e_t^{\text{Saber}} \mid x_t] < \mathbb{E}[e_t^{\text{Traditional}} \mid x_t]$.

\subsection{Proofs}
\label{app:proofs}

\subsubsection*{Proof of Lemma \ref{lem:aadu}}
\begin{proof}
Given the context $x_{t-1}$, the candidate set $\mathcal{D}_t$ and the threshold $\tau_t$ are deterministic. By the design of AADU, for every token index $i \in \mathcal{D}_t$, $c_i^t \ge \tau_t$ holds.

Based on Assumption \ref{ass:calibration} (specifically the upper bound property), the conditional probability of an error at position $i$ is bounded:
\begin{equation*}
\Pbb(\err_i \mid x_{t-1}) \le 1 - c_i^t + \delta_{\text{calib}}.
\end{equation*}

By the linearity of expectation, the conditional expected number of errors drafted within $\mathcal{D}_t$ is:
\begin{equation*}
\E[\err(\mathcal{D}_t) \mid x_{t-1}] = \sum_{i \in \mathcal{D}_t} \Pbb(\err_i \mid x_{t-1}).
\end{equation*}

Since $c_i^t \ge \tau_t \implies 1 - c_i^t \le 1 - \tau_t$ for all $i \in \mathcal{D}_t$, we can bound the expectation:
\begin{equation*}
\E[\err(\mathcal{D}_t) \mid x_{t-1}] \le \sum_{i \in \mathcal{D}_t} (1 - c_i^t + \delta_{\text{calib}}) \le \sum_{i \in \mathcal{D}_t} (1 - \tau_t + \delta_{\text{calib}}) = |\mathcal{D}_t|(1 - \tau_t + \delta_{\text{calib}}).
\end{equation*}
This establishes the conditional adaptive step-wise error injection bound, concluding the proof.
\end{proof}

\subsubsection*{Proof of Proposition \ref{prop:berm}}
\begin{proof}
By Assumption \ref{ass:calibration} (tight two-sided calibration), the conditional probability bounds for any historical token $j \in \mathcal{U}_{t-1}$ satisfy $\Pbb(\err_j \mid x_t) \le 1 - c_j^t + \delta_{\text{calib}}$ and $\Pbb(\err_j \mid x_{t-1}) \ge 1 - c_j^{t-1} - \delta_{\text{calib}}$. 
Thus, the relative conditional error growth for token $j$ is bounded by:
\begin{equation*}
\Pbb(\err_j \mid x_t) - \Pbb(\err_j \mid x_{t-1}) \le (1 - c_j^t) - (1 - c_j^{t-1}) + 2\delta_{\text{calib}} = \Delta_j + 2\delta_{\text{calib}}.
\end{equation*}

Therefore, $\Delta_j$ acts as a  valid proxy for the upper bound of the step-wise error degradation. The total upper bound of reliability degradation of the \textit{retained} context $\mathcal{U}_{t-1}^{\text{retain}} = \mathcal{U}_{t-1} \setminus \mathcal{R}_t$ is proportional to:
\begin{equation*}
\sum_{j \in \mathcal{U}_{t-1}} \Delta_j - \sum_{i \in \mathcal{R}_t} \Delta_i.
\end{equation*}

BERM partitions the historical context into the rollback set $\mathcal{R}_t$ (size $\mu_t$) and the retained set $\mathcal{U}_{t-1}^{\text{retain}}$. By algorithm design, BERM selects $\mathcal{R}_t$ such that the elements with the largest confidence drop ($\Delta_i$) are chosen. This guarantees that for any alternative pruning set $\mathcal{R}' \subseteq \mathcal{U}_{t-1}$ of size $\mu_t$:
\begin{equation*}
\sum_{i \in \mathcal{R}_t} \Delta_i \ge \sum_{k \in \mathcal{R}'} \Delta_k.
\end{equation*}

Because BERM explicitly maximizes the subtracted term $\sum_{i \in \mathcal{R}_t} \Delta_i$ via its greedy selection strategy, it minimizes the proxy upper bound of the remaining total error degradation within the retained context sequence given the pruning budget $\mu_t$. This completes the argument under the stated assumptions.
\end{proof}

\begin{figure*}[h!]
    \centering
    \includegraphics[width=0.97\textwidth]{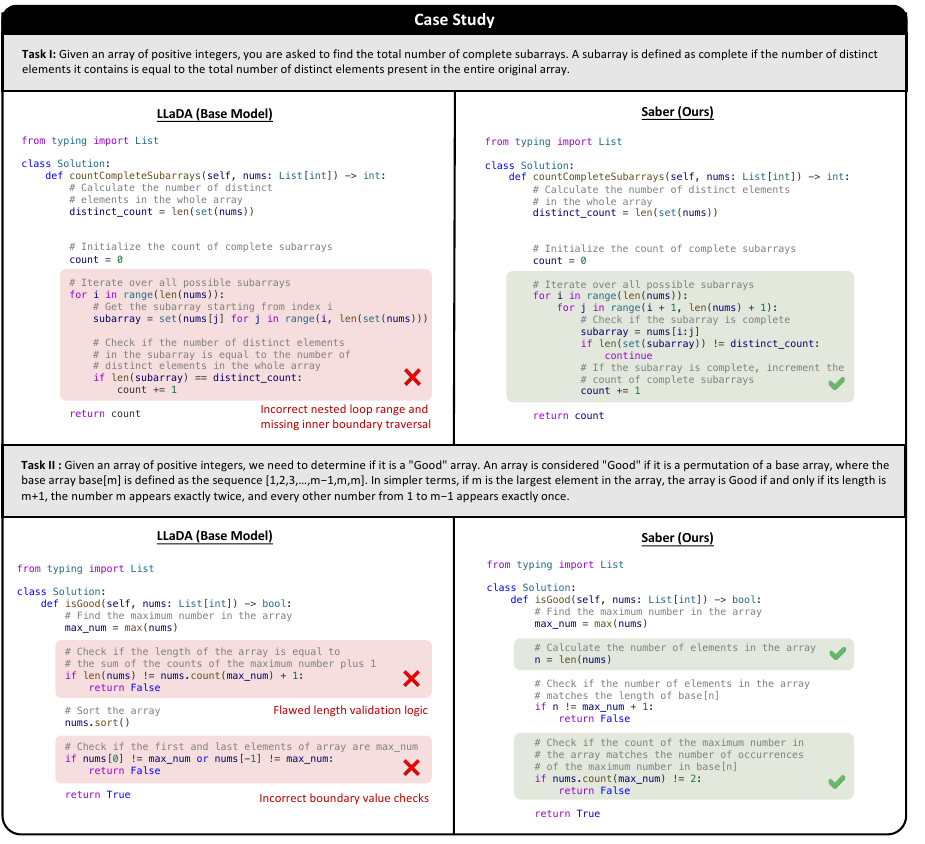}
    \caption{Case study.}
    \label{fig:qualitative}
\end{figure*}

\section{Case Study}

Figure \ref{fig:qualitative} presents a comparison of code generated by the base model LLaDA and Saber on two HumanEval problems. In Task I, LLaDA fails to enumerate all subarrays correctly. It uses a single loop with an incorrect range derived from the distinct element count, producing both logically flawed and syntactically invalid code. Saber instead generates the correct nested loop structure that systematically checks every subarray. In Task II, LLaDA misinterprets the ``Good array'' definition. Its length check relies on an incorrect relationship, and its boundary check after sorting is logically impossible for a valid sorted array. Saber correctly decomposes the problem into verifying the length condition (\textit{n=m+1n = m+1
n=m+1}) and the occurrence constraint (\textit{count(m)=2\text{count}(m) = 2
count(m)=2}). These examples illustrate how Saber's backtracking mechanism helps avoid the kind of early structural errors that the irreversible base sampler cannot recover from.
\\

\section{Computational Overhead Analysis}
\label{sec:computational_overhead}

In this section, we provide a detailed complexity breakdown and a comparative analysis of Saber's operations. Saber introduces two primary sources of per-step overhead, both of which are strictly bounded by the sequence length $N$:

\begin{itemize}
\item \textbf{Adaptive Acceleration via Dynamic Unmasking:} The process involves calculating a dynamic threshold (by averaging the confidence probabilities of previously unmasked tokens) and scanning masked positions to select candidates. These operations require only a single sequential pass, resulting in a time complexity of $\mathcal{O}(N)$.
\item \textbf{Backtracking-Enhanced Remasking Mechanism:} Identifying the lowest-confidence tokens to re-mask involves calculating confidence differences and finding the bottom-$\mu_t$ elements within the candidate set. This selection process can be executed efficiently in $\mathcal{O}(N)$ time using linear-time selection algorithms.
\end{itemize}

Crucially, these linear $\mathcal{O}(N)$ algorithmic operations are remarkably lightweight and are executed as CPU-based logic. In stark contrast, a single forward pass of a Discrete Diffusion Language Model (DLM) requires massive, heavily parameterized matrix multiplications on the GPU—such as self-attention mechanisms and feed-forward networks—which represent billions of operations and dominate the actual inference latency.

Empirically, our runtime breakdown confirms this theoretical analysis. The algorithmic overhead of Saber does not noticeably inflate the per-step execution time. Instead, because Saber's adaptive and self-correcting mechanisms effectively reduce the total number of required denoising iterations, the marginal per-step CPU overhead is overwhelmingly offset by the reduction in expensive GPU forward passes. Therefore, the overall computational cost is lower than that of standard sampling. As demonstrated in our main results (Table 1), Saber achieves an average overall inference speedup of $251.4\%$ under the condition of boosting Pass@1 accuracy by an average of
1.9\%. 

As a result, \textbf{the computational cost introduced by Saber in each step is entirely marginal, but its total computational cost is instead lower}, so it's highly acceptable for real-world code generation deployments.

\section{Evaluation with Different Configurations on other base DLMs}
\label{DifferentConfigurations}
In our main experiments, standard decoding settings (e.g., temperature $= 0$) were adopted to maintain consistency with previous DLM sampling literature. To comprehensively evaluate performance under different configurations, we reproduced the experiments following the best settings of Dream and DiffuCoder, where we directly followed the settings reported in their original paper for Dream and we conducted a grid search to determine the optimal settings for DiffuCoder (since its original paper did not report). As shown in Table \ref{tab:dream_settings}, Saber demonstrates stable and consistent improvements under these specific settings, validating the effectiveness of our approach across different configurations.

\begin{table}[ht]
\centering
\caption{Effectiveness of Saber compared to the baseline under the optimal settings of Dream and DiffuCoder on the HumanEval benchmark.}
\label{tab:dream_settings}
\begin{tabular}{llcccc}
\toprule
Base Model & Method & HumanEval $\uparrow$ & HumanEval-ET $\uparrow$ & Step $\downarrow$ & Time $\downarrow$ \\
\midrule
\multirow{2}{*}{Dream} & Baseline & 0.5749 & 0.4756 & 256 & 1:18:12 \\
 & Ours & \textbf{0.6037} & \textbf{0.5183} & \textbf{141.04} & \textbf{40:49} \\
\midrule
\multirow{2}{*}{DiffuCoder} & Baseline & 0.6829 & 0.5937 & 256 & 1:19:56 \\
 & Ours & \textbf{0.6890} & \textbf{0.6098} & \textbf{168.02} & \textbf{53:19} \\
\bottomrule
\end{tabular}
\end{table}

\section{Generalizability Across Different Domains}
\label{DifferentDomain}
To further evaluate the generalizability of Saber beyond code generation tasks, we extended our experiments to mathematical reasoning and general question-answering domains. Specifically, we evaluated our method against the default DLM sampling strategy on the MATH-500 and ARC-Challenge benchmarks. As shown in Table \ref{tab:generalization}, Saber consistently demonstrates superior performance and efficiency across both datasets. On MATH-500, Saber improves the Pass@1 score from 32.4\% to 34.0\% while reducing the total inference time by nearly 50\% and cutting the average decoding steps to 146. Similarly, on ARC-Challenge, Saber boosts accuracy from 53.51\% to 56.19\% with an even more substantial reduction in steps (from 256 to 129.66) and inference time. These results strongly confirm that the core mechanisms of Saber are not limited to tasks with strict structural constraints like code. Instead, Saber serves as a highly generalized, task-agnostic efficient sampling strategy for diffusion language models across diverse application scenarios.

\begin{table}[h]
\caption{Performance comparison on MATH-500 and ARC-Challenge benchmarks.}
\label{tab:generalization}
\centering
{
\begin{tabular}{lcccccc}
\toprule
\multirow{2}{*}{{Method}} & \multicolumn{3}{c}{{MATH-500}} & \multicolumn{3}{c}{{ARC-Challenge}} \\
\cmidrule(lr){2-4} \cmidrule(lr){5-7}
& {Pass@1 $\uparrow$} & {Step $\downarrow$} & {Time $\downarrow$} & {Pass@1 $\uparrow$} & {Step $\downarrow$} & {Time $\downarrow$} \\
\midrule
Baseline & 0.3240 & 256 & 3:01:02 & 0.5351 & 256 & 1:49:51 \\
Saber & \textbf{0.3400} & \textbf{146.00} & \textbf{1:31:15} & \textbf{0.5619} & \textbf{129.66} & \textbf{0:49:02} \\
\bottomrule
\end{tabular}
}
\end{table}

\section{Performance in Long-Context Scenarios}
To explore the sampling efficiency and correction behavior of diffusion language models under longer input conditions, we evaluated Saber in long-context scenarios on tasks with input lengths between 512 and 4096 tokens. Although long-context generation presents a significant challenge for DLMs, Saber demonstrates remarkable robustness, achieving more than three times the Pass rate of the baseline method, as shown in Table \ref{tab:long_context}.

\begin{table}[h]
\centering
\caption{Performance comparison in long-context scenarios (input lengths between 512 and 4096 tokens).}
\label{tab:long_context}
\begin{tabular}{lcc}
\toprule
Method & Performance & $\uparrow\Delta$ \\
\midrule
Baseline & 1.4\% & - \\
Saber (Ours) & \textbf{4.3\%} & \textbf{+207.2\%}\\
\bottomrule
\end{tabular}
\end{table}

\section{Hyperparameter Impact.}
We conduct a hyperparameter impact analysis of the backtracking ratio $\mu$ on the HumanEval benchmark to evaluate the robustness of our approach. As shown in Table~\ref{tab:hyperparameter}, Saber demonstrates strong robustness: the Pass@1 performance remains consistently high across a wide range of $\mu$ values (from $1/8$ to $1/2$). Crucially, all these configurations significantly outperform the baseline setting where no backtracking is applied ($\mu=0$, yielding a Pass@1 of 0.3523). This confirms that while the exact choice of $\mu$ offers a flexible trade-off between inference speed and accuracy, the backtracking mechanism itself provides a stable and substantial quality improvement regardless of minor hyperparameter variations.

\begin{table}[htbp]
\centering
\caption{Hyperparameter sensitivity analysis of the backtracking ratio $\mu$.}
\label{tab:hyperparameter}
\begin{tabular}{lccc}
\toprule
$\mu$ & Pass@1 $\uparrow$ & Steps $\downarrow$ & Time $\downarrow$ \\
\midrule
1/2  & 0.4512 & 158.40 & 59:12 \\
1/4  & 0.4451 & 128.55 & 48:54 \\
1/8  & 0.4512 & 118.92 & 41:55 \\
1/16 & 0.3963 & 109.93 & 39:59 \\
0 (w/o Backtracking-Enhanced Remasking) & 0.3523 & 65.67  & 28:30 \\
\bottomrule
\end{tabular}
\end{table}

\section{Extended Results of Generalizability across Different DLMs}
\label{differentbenchmark}
To further validate the generalizability of Saber, we conducted broader evaluation on the MBPP and MBPP-ET datasets in our default setting. As detailed in Table \ref{tab:broader_testing}, Saber achieves consistent improvements in Pass@1 scores, alongside significant reductions in decoding steps and inference time, across LLaDA, Dream, and DiffuCoder.

\begin{table}[h]
\centering
\caption{Broader testing of Saber on MBPP and MBPP-ET benchmarks across different DLMs.}
\label{tab:broader_testing}
\begin{tabular}{llcccc}
\toprule
Base Model & Method & MBPP $\uparrow$ & MBPP-ET $\uparrow$ & Step $\downarrow$ & Time $\downarrow$ \\
\midrule
\multirow{2}{*}{LLaDA} & Baseline & 0.4286 & 0.3138 & 256 & 3:12:08 \\
 & Ours & \textbf{0.4473} & \textbf{0.3302} & \textbf{110.96} & \textbf{1:33:33} \\
\midrule
\multirow{2}{*}{Dream} & Baseline & 0.6182 & 0.4637 & 256 & 2:38:03 \\
 & Ours & \textbf{0.6206} & \textbf{0.4660} & \textbf{131.48} & \textbf{1:26:16} \\
\midrule
\multirow{2}{*}{DiffuCoder} & Baseline & 0.5152 & 0.3677 & 256 & 2:42:28 \\
 & Ours & \textbf{0.5691} & \textbf{0.4192} & \textbf{125.65} & \textbf{1:20:51} \\
\bottomrule
\end{tabular}
\end{table}

\section{Extended Related Work}

\subsection{Code Generation}
Since the advent of artificial intelligence in the 1950s, code generation has been considered the Holy Grail of computer science research \cite{gulwani2017program}. With the rapid expansion of codebases and the increasing capacity of deep learning models, using deep learning for program generation has shown great potential and practicality \cite{RaychevVY14, LingBGHKWS16, Self-Collaboration, Agent4code, Self-Planning, ROCODE}. In recent years, the rise of pre-training techniques has brought new momentum to the field of code generation. For example, studies like CodeT5 \cite{CodeT5} and UniXcoder \cite{UniXcoder} pre-train models for code generation tasks. With the continual increase in model parameters, researchers have discovered emergent phenomena in LLMs, leading to new breakthroughs \nocite{GorM}. Against this backdrop, LLMs such as AlphaCode \cite{alphacode}, Codex \cite{codex}, Starcoder \cite{starcoder}, CodeLlama \cite{codellama}, and DeepSeek Coder \cite{DeepSeek_Coder} have emerged.

\section{Detailed Experimental Setup}
\label{app_setup}
In this section, we present the setups of our experiments below. 
\subsection{Datasets}
We conduct experiments on five code generation datasets to demonstrate the effectiveness of Saber, including HumanEval \cite{chen2021evaluating}, MBPP \cite{austin2021program}, HumanEval-ET and MBPP-ET \cite{CodeScore}, and LiveCodeBench \cite{jain2024livecodebench}. For all datasets, tasks are presented in a zero-shot format.

\begin{itemize}
    \item \textbf{HumanEval} is a widely used benchmark for evaluating LLMs’ ability to generate correct Python functions from docstrings.
    
    \item \textbf{MBPP} (Mostly Basic Python Problems) consists of small-to-medium Python programming tasks designed to test basic algorithmic reasoning.
    
    \item \textbf{LiveCodeBench} is a contamination-aware benchmark that continuously collects new programming problems from contest platforms (LeetCode, AtCoder, Codeforces) and focuses beyond simple code generation to broader code reasoning capabilities.
    
    \item \textbf{HumanEval-ET and MBPP-ET} are extended versions of the original HumanEval and MBPP. They augment each task with over 100 additional test cases and include edge-case tests, which enhance the reliability of the evaluation.
\end{itemize}

\subsection{Baselines}
We conduct a comprehensive evaluation of Saber against established baseline decoding methods for DLMs. The results confirm that Saber achieves superior performance, effectively validating its effectiveness.

\begin{itemize}
    \item \textbf{Standard DLM sampling (Default)}:In this mode, DLM generates responses by continuously decoding over a predetermined full output length. The decoding methods include confidence-based, entropy-based, and random approaches.
    
    \item \textbf{Efficient DLM sampling Methods}, including: \textbf{Semi-autoregressive (SAR) \cite{llada}}: This strategy decodes in blocks from left to right. It thus combines aspects of autoregressive order with diffusion’s simultaneous updates. Within each block, tokens are decoded based on confidence; \textbf{parallelism increase (p)}: increasing the number of tokens per sampling; \textbf{WINO} \cite{hong2025wide}: using a fixed threshold for acceleration; \textbf{Fast-dLLM} \cite{Wu2025FastdLLMTA}: using cache for acceleration; and \textbf{ReMDM} \cite{remdm}: using a fixed threshold for remasking.

    \item \textbf{Parallelism Increase (p)}, \textbf{Semi-autoregressive (SAR)} \cite{llada}, \textbf{WINO \cite{hong2025wide}, Fast-dLLM \cite{Wu2025FastdLLMTA}}, and \textbf{ReMDM} \cite{remdm} are recently proposed efficient DLM sampling methods.
\end{itemize}

\subsection{Metric}
Our evaluation employs \textbf{Pass@1} as the primary metric. It is calculated as the percentage of problems for which the generated code passes all test cases with a single attempt. The formula is as follows:
$$ \text{pass}@1 = \frac{1}{|N|} \sum_{i=1}^{|N|} \mathbb{I}(\text{Passed}(\text{Generation}_i)) $$
where $|N|$ is the total number of problems, and the indicator function $\mathbb{I}(\cdot)$ is 1 if the single generation for a given problem passes all its test cases, and 0 otherwise. 

In addition to performance, we also measure the \textbf{Step} (i.e., average generation steps per sample) and \textbf{Time} (i.e., total generation time).

\subsection{Implementation Details}
In this paper, we employ the LLaDA-8B-Instruct \cite{llada} as the base model for our experiments. The default temperature for all baselines is set at 0 for consistency. In the fixed-length decoding scenario, we set the generation length to 256 tokens. For the semi-autoregressive approach, the block length was configured to 128. All other efficient DLM sampling methods follow the same configuration as their original paper. To mitigate the instability of model sampling, we report the average results of five trials in the experiments. All experiments were conducted on a workstation equipped with 8 NVIDIA A6000 GPUs (48GB each) and 1TB RAM, with a single GPU for each experiment.

\section{Limitation}
Our work has the following two main limitations. 

First, Saber requires slightly more computational resources than direct sampling in a DLM sampling step. 
Specifically, Saber requires $O(N)$ CPU-based operations (where $N$ is the sequence length) to execute the dynamic unmasking and backtracking remasking logic. However, compared to the massive GPU-intensive matrix multiplications required by DLM forward passes, it is marginal and acceptable. Furthermore, since Saber significantly reduces the total number of denoising iterations, the overall computational cost is actually lower than that of standard sampling.
Detailed discussion can be found in Appendix \ref{sec:computational_overhead}.

Second, we only explore the choice of hyperparameters within reasonable ranges, considering the trade-off between performance and speed, as shown in Table~\ref{tab:hyperparameter} and the right of Figure~\ref{fig:intro}. 
Further optimization of hyperparameters could yield additional improvements.

\end{document}